%% file: main.tex
\pdfoutput=1
\documentclass[11pt]{article}

\input{sections/header}

\begin{document}
\maketitle

\input{sections/0_abstract}
\input{sections/1_introduction}
\input{sections/2_related_work}

\input{sections/3_method}

\input{sections/4_experiment}

\input{sections/5_conclusion}
\input{sections/6_ethics}

%\appendix
\bibliographystyle{acl_natbib}
\bibliography{sections/custom}

\end{document}

%% file: sections/header.tex
\usepackage[]{emnlp2021}

\usepackage{times}
\usepackage{latexsym}
\usepackage{enumitem}

\usepackage[T1]{fontenc}
\usepackage[utf8]{inputenc}

\usepackage{microtype}
\usepackage{color}
\usepackage{booktabs}
\usepackage{graphicx}
\usepackage{makecell}
\usepackage{amsmath}
\usepackage{tikz}
\usetikzlibrary{plotmarks}

\usepackage{xcolor}
\usepackage{multirow}

\usepackage{cleveref}
\crefformat{section}{\S#2#1#3}
\crefformat{subsection}{\S#2#1#3}
\crefformat{subsubsection}{\S#2#1#3}
\crefrangeformat{section}{\S\S#3#1#4 to~#5#2#6}
\crefmultiformat{section}{\S\S#2#1#3}{ and~#2#1#3}{, #2#1#3}{ and~#2#1#3}
\usepackage{refstyle}
\Crefformat{figure}{#2Fig.~#1#3}
\Crefmultiformat{figure}{Figs.~#2#1#3}{ and~#2#1#3}{, #2#1#3}{ and~#2#1#3}
\Crefformat{table}{#2Tab.~#1#3}
\Crefmultiformat{table}{Tabs.~#2#1#3}{ and~#2#1#3}{, #2#1#3}{ and~#2#1#3}
\Crefformat{appendix}{\S#2#1#3}
\crefformat{algorithm}{Alg.~#2#1#3}

\newcommand{\stitle}[1]{\vspace{0.3em} \noindent{\bf #1.}}

\title{Table-based Fact Verification with Salience-aware Learning}

\author{Fei Wang,\;~Kexuan Sun,\;~Jay Pujara,\;~Pedro Szekely \and Muhao Chen \\
  Department of Computer Science \& Information Sciences Institute\\
  University of Southern California \\
  \texttt{\{fwang598,kexuansu,jpujara,szekely,muhaoche\}@usc.edu} 
  }

\date{}

%% file: sections/0_abstract.tex
\begin{abstract}

Tables provide valuable knowledge that can be used to verify textual statements.
%Existing works show that understanding the connections between the table and the statement is important to table-based fact verification.
%However, there is no supervision signal for learning the connections directly.
While a number of works have considered table-based fact verification, direct alignments of tabular data with tokens in textual statements are rarely available.
Moreover, training a generalized fact verification model requires abundant labeled training data.
% Previous works only explored template-based data augmentation ignoring that statements can be presented in very diverse ways.
In this paper, we propose a novel system to address these problems.
Inspired by counterfactual causality, our system identifies token-level salience in the statement with \textit{probing-based salience estimation}. 
%We leverage salience-aware learning from two perspectives.
Salience estimation allows enhanced learning of fact verification from two perspectives.
From one perspective, our system conducts \textit{masked salient token prediction} to enhance the model for alignment and reasoning between the table and the statement.
From the other perspective, our system applies \textit{salience-aware data augmentation} to generate a more diverse set of training instances by replacing non-salient terms. 
Experimental results on TabFact show the effective improvement by the proposed salience-aware learning techniques, 
% leads to significant improvements over the current SOTA systems.
% significantly outperforms previous SOTA systems.Detailed analysis demonstrates the gain obtained by the proposed techniques.
leading to the new SOTA performance on the benchmark.
\footnote{Our code is publicly available at \url{https://github.com/luka-group/Salience-aware-Learning}}

\end{abstract}

%% file: sections/1_introduction.tex
\section{Introduction}

% Paragraph 1: Fact/statement verification is important; Tables are reliable sources of evidence
% Paragraph 2: A common solution involve a process of (i) retrieving relevant and important evidence (ii) infer the logical consequence between the evidence(s) and the statement. Common challenges to existing methods: (i) identifying salience of facts in tables; (ii) lack of learning resources?
% Paragraph 3: 

% Flow of paragraph 1: 1. Fact verification is important; (2. Textual resources in earlier studies may be noisy;) 3. Recent works use Tables are reliable sources of evidence; 4. Any downstream applications?
%Fact verification is an essential research topic in the age where misinformation appears everywhere on the Internet and is hard to be distinguished
%With the rapid growth of the internet, we are living in an era with both information explosion and information pollution \cite{zhang2019evidence}.
%Hence, in natural language understanding, fact verification has become a particularly important %natural language understanding 
Fact verification, the problem of determining whether a statement is entailed or refuted by evidence, has quickly become a critical problem in NLP to combat information pollution~\cite{rashkin2017truth,thorne2018fever,zhang2019evidence, zellers2019defending, wadden2020fact}. 
Successful fact verification enables downstream tasks such as misinformation detection, fake news identification, factual error correction, and deceptive opinion detection \cite{ott2011finding, shu2017fake, yoon2019detecting, cao2020factual}.
%systems, and can identify fake news 
%factual error correction \cite{
% incongruent news headline detection \cite{yoon2019detecting} 
%and deceptive opinion detection \cite{

\begin{figure}[t]
\centering
\includegraphics[scale=0.42]{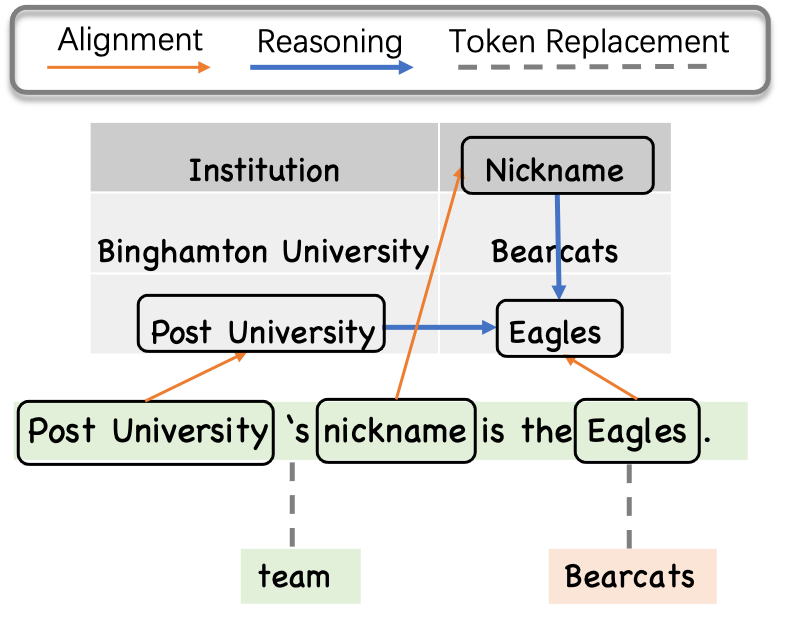}
%\vspace{-0.5em}
\caption{
An example of table-based fact verification, 
% where the entailed statements are colored green and the refuted statement is colored red.
with green for entailed statements and red for refuted statements.
% Masking and predicting significant words (e.g. \textit{'Eagles'}) in a statement will urge the model to learn the essential abilities to this task (i.e. alignment and reasoning).
Alignment and reasoning are essential for both table-based fact verification and masked salient token prediction (e.g. \texttt{''Eagles''}).
Token replacement may lead to similar (e.g. \texttt{'''s''} to \texttt{''team''}) or different (e.g. \texttt{''Eagles''} to \texttt{''Bearcats''}) statements.
% The figure also shows the alignments between the first statement and the table.
% the matching between substrings of the first statement and substructures of the table. 
% The alignment and reasoning can be more difficult when the logical operations in the statement are more complex (e.g. sum) or the statement consists of several substatements.
% The fine-grained connections between the first statement and the table is marked with rectangle. 
% Note that the words in statements and the cells in tables are connected when they are semantically consistent and may not match in terms of string.
}
\label{fig:intro}
%\vspace{-1em}
\end{figure}

% Main problem and existing works
Recently, table-based fact verification \cite{chen2019tabfact, zhong2020logicalfactchecker, yang2020program} has garnered attention. 
As a ubiquitous and clean format of semi-structured knowledge, tables are regarded as reliable sources of evidence to verify the textual statements \cite{chen2019tabfact}.
%As an ubiquitous source of reliable information, the tabular data are introduced as evidence for fact verification recently .
Leveraging tabular data for fact verification requires identifying relevant evidence in tables,
% and inferring the logical relation between the structured evidence and the statement.
and conducting logical reasoning according to the selected evidence.
Prior studies have attempted to 
generate logical programs to capture logical operations and relations between the statement and the table \cite{zhong2020logicalfactchecker,yang2020program, shi2020learn}.
% and adopt Transformer-based language model to benefit from general as well as task-specific pre-training without counting on explicit logical programs \cite{eisenschlos2020understanding}.
More recent work shows that Transformer-based language models with general and task-specific pre-training over textual and tabular data can achieve SOTA performance without counting on explicit logical programs \cite{eisenschlos2020understanding, dong2021structural}.

% To do so, earlier studies have attempted to generate logical programs to capture the relation with symbolic operations 
% % (e.g. \textit{count} and \textit{aggregation})
% \cite{zhong2020logicalfactchecker,yang2020program}. 
% In order to benefit with automated feature representation and %discourse relation inference
% natural language inference (NLI), more recent studies propose to develop table-based fact verification methods on the basis of language models \cite{chen2019tabfact, eisenschlos2020understanding}.
% Through this line of research, one of the SOTA systems \cite{eisenschlos2020understanding} adopts
% TAPAS \cite{herzig2020tapas}, a Transformer-based language model pre-trained on both textual and tabular data, to table entailment
% without counting on explicit logical programs. 

% The learning process of logical program generation is guided by the verification results.
%, as there is no label for semantic parsing. 
% In order to leverage pretraining
% In order to gain some additional benefits from more recented advances using pre-training,
% In order to gain some additional benefits from recent advances in language pre-training,
% , and achieves SOTA performance 
% a more recent system \cite{eisenschlos2020understanding} improves the Transformer-based language model pre-trained on both textual and tabular data, TAPAS \cite{herzig2020tapas}, to pre-train on entailment tasks, and achieves SOTA performance without counting on explicit logical programs. 

% Common challenge to existing methods
% The challenge should be directly related to the task.
However, %some important problems remain unsolved.
%there are several challenges of the table-based fact verification task that have still been overlooked.
to provide a reliable solution to the table-based fact verification task, several critical challenges are still overlooked by prior studies.
% One challenge is to effectively bridge salient evidences in tables to salient words in statements. 
One challenge is to effectively provide connections among components of the statement and substructures of the table, and accordingly conduct the inference.
%\muhao{Why this is naturally a benefit? need one sentence}
% One challenge is to effectively identify the fine-grained connections between the statement and the table and conduct logical reasoning.
Being unaware of such fine-grained connections and logical relations could raise the risk of misalignment, incorrect reasoning and ignoring salient components of a statement, % which are entailed or refuted by the evidence in a table,
and therefore leads to incorrect verification results.
For example, to verify the statement in \Cref{fig:intro}, the model should implicitly or explicitly infer all the five arrows accurately.
% \muhao{maybe refer to the example in Figure 1.}
% Being unaware of the exact connections between them will let models make same predictions for incomplete or opposite statements where some salient words are absent or replaced.
Although some works have tried to perform token-level interactions and generate logical programs to connect statements and tables and conduct logical reasoning \cite{zhong2020logicalfactchecker,yang2020program}, the supervision signals to guide the learning process are typically sparse. % for learning the connections.
% \todo{examples} % bad case for previous models
Another challenge is that training a well-generalized fact verification model %is resource-hungry.
non-trivially requires abundant labeled training data.
% learning the verification model is resource hungry. 
% The same meaning can be expressed differently.
% For example, ...
Limited training data can only cover limited statement patterns and hinder robustness and generalizability of model inference.
Previous works either trained on limited data \cite{zhong2020logicalfactchecker,yang2020program} or %augment new data by replacing entities or filling a few synthetic phrases \cite{eisenschlos2020understanding}. 
augment training data with specific statement generation templates \cite{eisenschlos2020understanding}.
% using rule-based generation methods with a small vocabulary \cite{eisenschlos2020understanding}.
%The insignificant words in statements, which are also important to robustness, are ignored.
Yet, in real-world scenarios, statements and evidences can be presented in very diverse ways, and such diversity is difficult to be comprehensively captured by specific templates.
% Previous models trained on limited data may react inconsistently to statements with similar meanings.
% When learned with limited training data, the robustness of prior models can be hindered, causing them to react inconsistently to statements with similar meanings.
% lack of learning resources which hi robust to statements of the same meaning with tiny differences. 
% \todo{examples} 
% Figure show that previous models are not robust to statements of the same meaning with tiny differences. 

% First, previous works do not have direct supervision signals for bridging salient evidences in tables to salient words in statements. 
% As a consequence, these methods may make the same predictions for incomplete or opposite statements where some salient words are absent or replaced.
% \todo{Figure 1a shows that the both Symbolic and TAPAS model make the same prediction when an important word in the statement is replaced by other words (or masked).
% This phenomenon indicates that the TAPAS model is not aware of the relationship between the salient words in statements and corresponding salient evidence in tables.}
% \todo{Figure 1b shows that the predictions by the TAPAS model are not consistent when we replace the non-salient word in a statement by other valid words. 
% This phenomenon indicates that the TAPAS model is not robust to the non-salient words.}
% can be fooled by replacing non-salient words.

%In this paper, we address the aforementioned problems by leveraging multi-task learning and data augmentation.
To this end, we propose a novel salience-aware learning system for table-based fact verification.
%by leveraging multi-task learning and data augmentation.
% with salience-aware learning and %more generalized automated data augmentation strategy.
% probabilistic data augmentation.
Starting from a TAPAS \cite{herzig2020tapas} language model fine-tuned on the TabFact dataset, 
our system identifies salient and non-salient tokens in statements with a \emph{probing-based salience estimation} method inspired by counterfactual causality \cite{pearl2009causality} (\Cref{sec:salient}).
Then, the system leverages the estimated salience information from two perspectives.
From one perspective, to enhance the model for capturing fine-grained connections and supporting the reasoning between statements and tables, the system conducts \emph{masked salient token prediction} as an auxiliary task (\Cref{sec:token_prediction}). 
% By reusing the embedding layer of TAPAS as a language model head and training the model to fill in the blanks of salient tokens in entailed statements, the model can directly learn to assign values in tables to the mask position in statements.
% More specifically, this task is to reuse the embedding layer of TAPAS as a language model head and train the model to predict the masked salient token in an entailed statement given the corresponding table.
More specifically, this task is to predict the masked salient token in an entailed statement given the corresponding table by reusing the embedding layer of TAPAS as a language model head.
The fact verification task can %benefit from
receive indirect supervision from the auxiliary task, as both of them requires table-text alignment and logical reasoning. 
% Since all the parameters of the masked language model are from the TAPAS-based fact verification model, it can benefit from the auxiliary task directly. 
From the other perspective,
to improve the model robustness, %we leverage 
instead of using templates for statement augmentation like prior work \cite{eisenschlos2020understanding},
we develop a \emph{salience-aware data augmentation} technique %as a data augmentation method 
(\Cref{sec:aug}). 
Intuitively, replacing non-salient tokens % seeks to 
provides unseen statements while preserving the meaning and correctness of the original statement.
% Accordingly, 
This strategy enhances the size and comprehensiveness of the training data and %improves robustness and generalizability of the model.
further complements training with more supervision signals.
% why augmentation ... 
% We compare our proposed method with previous works on the TabFact dataset. 

The main contributions of this paper are three-fold.
First, we propose a probing-based salience estimation method to evaluate the importance of each token in a statement according to the counterfactual causality theory. 
Second, we propose a novel salience-aware learning system that helps the fact verification model to find the connections between %able-statement alignment
the table and the statement, %and logical reasoning in terms of salient tokens and model robustness to diverse statement patterns driven by non-salient tokens.
and enhance the inference ability of the model with the auxiliary task of masked salient token prediction. %and data augmentation driven by non-salient tokens.
Third, to complement with insufficient training signals and improve the model robustness on %diverse statement patterns,
heterogeneous statements, we incorporate a probabilistic data augmentation method driven by non-salient tokens.
%Third, our system outperforms the previous SOTA method on the TabFact benchmark, and achieves better performance on both simple and complex statements.
We evaluate our system based on the TabFact benchmark, which shows promising performance on this task and drastically outperforms prior methods.
Detailed analysis demonstrates the effectiveness and essentially of both masked salient token prediction and salience-aware data augmentation techniques for the improved performance.

% However, while performing well, directly applying pre-trained language models to table-based fact verification raises the following problem: \textit{Does the model make decisions based on semantic relations between tables and statements, or based on some unrelated words in them?} 
% To be more specific, there is no guarantee that the pre-trained language models are aware of salient facts in statements, and are robust to non-salient words. (Figure shows that TAPAS may make same prediction without salient word & different prediction without non-salient token)

% Know what you should know: select and predict key facts in table based fact verification.

% 1. Joint MLM (why and how)
% 2. Select token by counterfactual (Importance = P(y|c+token;M) - P(y|c+mask;M))
% (optional) 3. mask salient tokens to get sentence templates. Then let the MLM model fill in each template. each time fill one token.  

% The relationship between MLM and fact verification:
% enumerate all possible tokens + fact verification -> solve MLM
% MLM is an ability fact verifier must have
% MLM has denser supervision signals
% https://cstheory.stackexchange.com/questions/32946/for-which-problems-in-p-is-it-easier-to-verify-the-result-than-to-find-it

% Select token by counterfactual
% Assumption:
% statement is the event, label is the consequence, token to mask is a factor

\begin{figure*}[t]
\centering
\includegraphics[clip, trim=0cm 10cm 10cm 0.5cm, width=1.00\textwidth, scale=0.38]{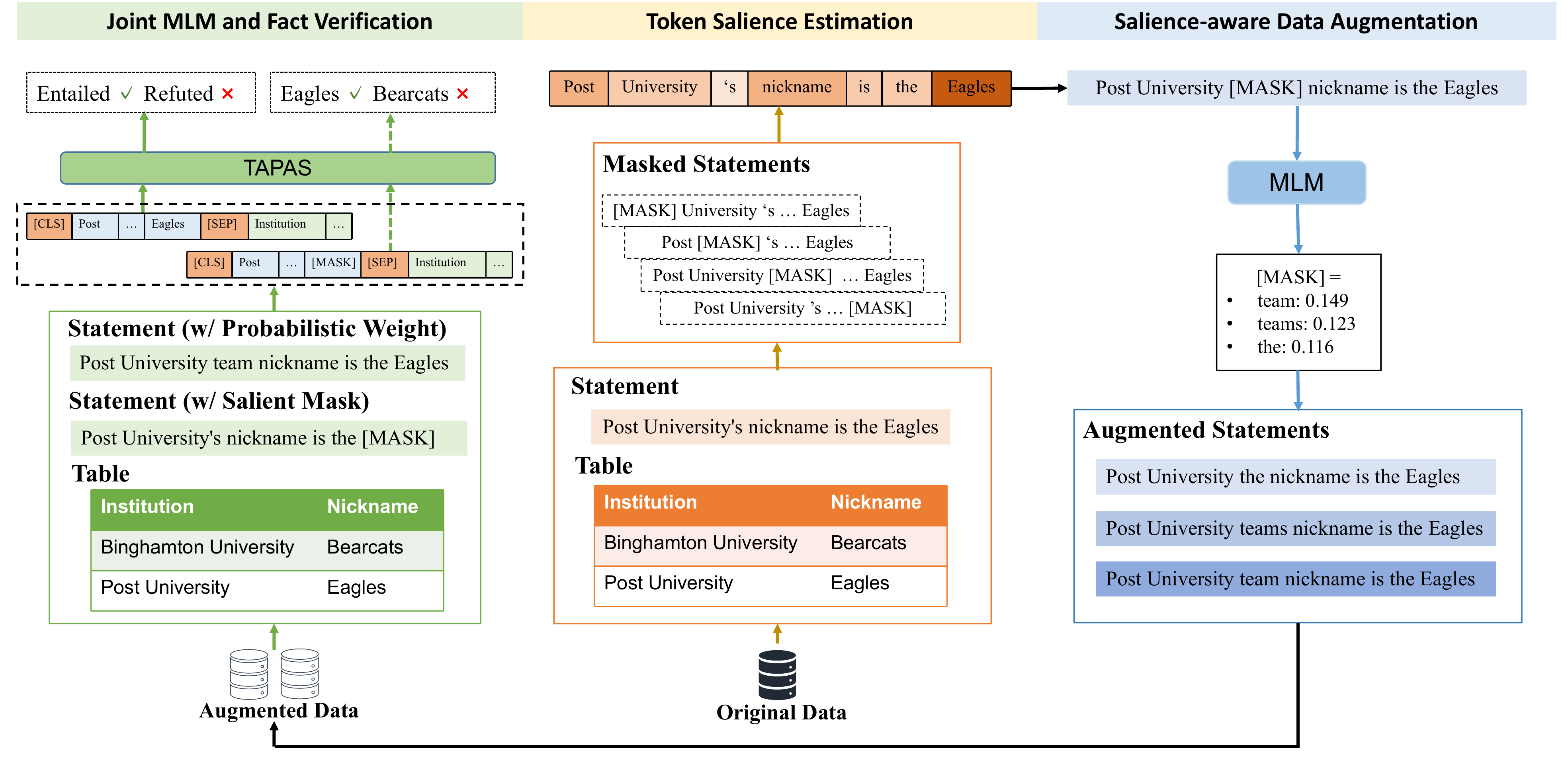}
%\vspace{-0.5em}
\caption{
Workflow of the proposed system. The system is composed of three parts. The arrows illustrate how information is transferred. For tokens, a lighter background color indicates a lower salience score. For augmented statements, a lighter background color indicates a smaller probability.
% \todo{MLM and token salience estimation also input the table. MLM model is also TAPAS. We'd better find another term for joint MLM and fact verification.}
}
\label{fig:system}
%\vspace{-1em}
\end{figure*}

%% file: sections/2_related_work.tex
\section{Related Work}

% We provide a selected summary for two topics.
In this section, we provide a selected summary for two related research topics.

\subsection{Fact Verification}
% FEVER -> TABFACT
Fact verification have become an essential research topic in recent years with the rising concerns of misinformation \cite{vlachos2014fact, wang2017liar, thorne2018fever, khattar2019mvae, zellers2019defending, chen2019tabfact}. 
Early works on fact verification are mainly based on unstructured textual evidence \cite{yin2018twowingos, nie2019combining, zhou2019gear}. 
% These works focus on both evidence retrieval and claim verification according to the selected evidence.

% table-based fact verification
Recently, much attention has been paid to table-based fact verification \cite{chen2019tabfact, zhong2020logicalfactchecker, yang2020program, eisenschlos2020understanding, shi2020learn, dong2021structural}. \citet{chen2019tabfact} released the TabFact benchmark, %and proposed two baselines based on a language model and lexical matching, respectively.
%TabFact
and %has 
motivated two lines of research. 
% first line
Considering the importance of logical operations in this task, some works introduce such inductive bias by explicitly generating and capturing logical programs.
% Latent Program Algorithm (LPA) \cite{chen2019tabfact} is a logical program-driven baseline. 
Latent Program Algorithm (LPA) \cite{chen2019tabfact} collected potential program candidates and execution results according to a search algorithm, and then trained a Transformer-based \cite{vaswani2017attention} model to assign a confidence score to each program %using weakly supervision signals.
based on matching to the statement.
Through this line, later works have explored improved ways to generate and capture logical programs \cite{zhong2020logicalfactchecker, yang2020program}.
LogicalFactChecker \cite{zhong2020logicalfactchecker} generated logical programs using a sequence-to-action generation approach, %motivated by semantic parsing. The system 
where it applied neural module networks \cite{andreas2016learning} to capture the logical structure of programs.
HeterTFV \cite{shi2020learn} learned to combine linguistic information and symbolic information with a heterogeneous graph attention network.
ProgVGAT \cite{yang2020program} verbalized the execution processes of the generated programs, and applied graph attention networks \cite{velivckovic2017graph} to capture each execution tree.
% second line
Beside logical programs, other studies applied pre-trained language models to linearized tables and perform fact verification as natural language inference (NLI) \cite{chen2019tabfact, eisenschlos2020understanding, dong2021structural}.
Table-BERT \cite{chen2019tabfact} applied BERT \cite{devlin2019bert} as an NLI model. 
\citet{eisenschlos2020understanding} and \citet{dong2021structural} improve this strategy by conducting task-specific pre-training to TAPAS \cite{herzig2020tapas}, a Transformer-based language model pre-trained on both textual and tabular data.
% misc
% In addition, \citet{schlichtkrull2020joint} extended the TabFact dataset to joint learning for evidence retrieval and fact verification.

% Our work is connected to the second line of table-based fact verification, applying pre-trained language models on linearized tables. Different from theirs, our proposed method leverages multi-task learning introducing direct supervision signals to support fine-grained statement-table interaction and reasoning, and applies data augmentation to improve model robustness to insignificant words. 
Our work takes advantages of both lines of research on table-based fact verification, introducing cross-structural alignment bias and logical reasoning bias to pre-trained language models. 
Besides, previous works focus on significant words in statements, while we apply data augmentation to improve model robustness to insignificant words.

\subsection{Counterfactual Causality in NLP}
Counterfactual thinking and causal inference have inspired several studies in natural language processing, including
counterfactual story rewriting \cite{qin2019counterfactual},
paraphrasing diversification \cite{park2019paraphrase},
measuring fairness in text classification \cite{garg2019counterfactual},
debiasing in machine translation \cite{saunders2020reducing} and visual question answering \cite{niu2020counterfactual}.
This direction has also developed data augmentation strategies in various NLP tasks \cite{zmigrod2019counterfactual, kaushik2019learning, fu2020counterfactual, zeng2020counterfactual}.
% introducing methods and resources for training models less sensitive to spurious patterns in sentiment analysis and natural language inference \cite{kaushik2019learning} (avoid spurious correlation)
% zeng2020counterfactual NER
Especially, counterfactual causality has been used to measure the causal effects of specific inputs in visual question answering \cite{niu2020counterfactual}.

Inspired by these applications, we apply the thought of counterfactual causality on table-based fact verification, and detect token-level salience in statements in a probing manner.

% \subsection{Multi-task Learning in NLP}

% \subsection{Data Augmentation in NLP}

%% file: sections/3_method.tex
\section{Method}
In this section, we describe the technical details of the proposed system. 
% follows the NLI formulation of table-based fact verification with a finetuned TAPAS language model (\Cref{sec:basic_model}).
% Our system is based on a multi-task learning process which learns to conduct fact verification as the main task (\Cref{sec:basic_model}) and masked salient token prediction as an auxiliary task (\Cref{sec:salient}-\Cref{sec:token_prediction}). 
Our system extends the NLI formulation of table-based fact verification \cite{eisenschlos2020understanding} with the pretrained language model TAPAS as the backbone (\Cref{sec:basic_model}).
As a preliminary step, our system estimates token-level salience in a probing manner %and identifies the most salient and non-salient tokens in
for each statement (\Cref{sec:salient}). 
% In addition, %a dedicated masked language model (MLM) is used to predict tokens as a way to augment training data \Cref{sec:aug}.
% we leverage the estimated token salience to incorporate a probabilistic data augmentation (\Cref{sec:aug}), seeking to enhance the verification model with more automatically induced supervision signals in training (\Cref{sec:training}).
The proposed salience-aware learning %method includes two techniques.
leverages the estimated salience information from two perspectives.
From one perspective, it enhances the main task learning with an auxiliary task of masked salient token prediction (\Cref{sec:token_prediction}). 
In this auxiliary task, our system masks salient tokens in entailed statements and requires the model to jointly solve the cloze task along with the main task of fact verification.
From the other perspective, our system incorporates a probablistic data augmentation technique (\Cref{sec:aug}) by replacing non-salient tokens in statements according to a pretrained masked language model (MLM). 
This is followed by the technical details of training and inference processes (\Cref{sec:training}).
The overall architecture of our system is shown in \Cref{fig:system}.

\subsection{Base Model for Fact Verification}\label{sec:basic_model}
% TAPAS for fact verification

Our system %is an extension of the previous SOTA method 
adopts the TAPAS \cite{herzig2020tapas} model from the previous SOTA method as the base model.
%\cite{eisenschlos2020understanding}, which formulates table-based fact verification as a NLI task and applies TAPAS \cite{herzig2020tapas} to solve it.
In this way, we also formulate the main task of table-based fact verification as an NLI task following \citet{eisenschlos2020understanding}.

% In the multi-task learning process, we treat the fact verification task as the main task. One of the current SOTA methods~\cite{eisenschlos2020understanding} that address %the table-based fact verification
% this task %follows the architecture of a question-answering model TAPAS because of the effectiveness it has shown in representing both tabular and textual data. 
% reveals the superiority of the TAPAS language model due to its effectiveness in representing both tabular and textual data. 
% To be specific, 
% TAPAS extends BERT with six types of additional positional embeddings: token embeddings, position embeddings, segment embeddings, column and row embeddings, and rank embeddings.
For a brief description of TAPAS,
it extends BERT’s architecture \cite{devlin2019bert} with additional positional embeddings to represent tabular structure.
Specifically, in addition to the embeddings used by BERT, the model applies column and row embeddings to represent the column index and row index of the cell enclosing the token, and rank embeddings to represent the numeric rank of the cell referring to the token if the column is sortable.
% To combine both tabular and textual data, 
It flattens %a table
% table cells sequentially into a list of tokens, and concatenates tokens from tables and tokens from surrounding text. 
the table into a sequence of words and concatenates them with textual sequence if any as input.
% During pretraining, two learning objectives are used. 
% One is the MLM objective, %which has been commonly used for pre-training language models, 
% and other is an alignment objective to predict if a table belongs to the text. 
The model is pre-trained using an MLM objective.
% For the MLM objective, it uses whole word masking for the text and whole cell masking for the table.
\citet{eisenschlos2020understanding} designed task-specific intermediate pretraining tasks to improve the model performance on table-based fact verification.
We use the model released by them as our basic model.
Following their setting, we add a \texttt{[CLS]} token at the beginning of the input sequence, and separate the statement and the linearized table with a \texttt{[SEP]} token.
Then, our system adopts the TAPAS model to encode the input sequence and model the probability of entailment with a task-specific prediction head taking the final representation of the \texttt{[CLS]} token as input.
Specifically, the task-specific prediction head is implemented as an MLP with the sigmoid activation fuction for binary classification, which is consistent with \citet{eisenschlos2020understanding}.

% In our method, we use TAPAS as an encoder, on top of which a multi-layer perceptron is added as a classification layer. Given an evidence sequence $E_t$ and a statenent $E_e$, we model the probability of entailment as
% \[
%     P(s|T) = MLP(E_t|E_e)
% \]
% \todo{this is similar to the Understanding xxx paper. We may need to change}

% \subsection{Salient Token Detection}

\subsection{Probing-based Salience Estimation}\label{sec:salient}
% Counterfactual Causality
% salient: values in table; aggregated values; words indicating logical relations
Lexical tokens usually have different levels of %task-specific 
importance with regard to the overall content or purpose of a description \cite{chiarcos2011introduction, liu2018automatic, xiong2018towards}.
% \cite{rennie2005using, kireyev2009semantic, wu2013measuring}. 
For example, in the sentence ``\texttt{Post University
has used the Eagles as its nickname}'', the tokens like ``\texttt{Eagles}'' and ``\texttt{nickname}'' are more important than others such as ``\texttt{has used}'' and ``\texttt{as}'' for determining if the sentence is refuted or entailed. 
We refer to such highly important tokens as \textit{salient} tokens, and less important ones as \textit{non-salient} tokens. 
To make use of token-level salience in the table-based fact verification task, the %first
immediate challenge is to \textit{estimate the salience of each token in a statement}.

%We address the challenge by leveraging counterfactual theories of causation. 
Inspired by the counterfactual theories of causation \cite{pearl2009causality, lewis2013counterfactuals}, we address the challenge with a probing-based salience estimation method.
Counterfactual causality has been widely used in social science for measuring the causal effects of specific factors \cite{tetlock1997counterfactual, bradycausation, morgan2015counterfactuals}, and has also been introduced to deep learning \cite{tang2020unbiased, niu2020counterfactual}.
In %terms of table-based fact verification
our context of fact verification, the intuition of counterfactual causation is to testify that: 
\textit{If the model has not seen the token, will it still make the same prediction?}
The counterfactual lies between the fact that the token is seen and the imagination that the token is masked.
The comparison between them naturally reflects the effect of the token, because the token is the only thing changed between the two situations.
% In other words, 
% given a token, %if the model makes different predictions with and without seeing the token, the token is salient.

Technically, to estimate the salience of a token in a statement, 
% we probe to see if removing the token will cause the fact verification model to make drastically different prediction.
we compare the confidence score to the gold fact verification label between the statements with that token unmasked and masked.
% If so, then this token will be considered salient.
% \muhao{It feels like from here, the rest part is still under development.}
Formally, given the table $T$, original statement $S$ and its counterfactual version $S^\prime_{t}$ with the target token $t$ masked, the salience score of $t$ in this statement is
 \[
 salience(t) = \bigg| P(y|S, T) - P(y|S^\prime_{t}, T) \bigg|
 \]
where $y$ indicates the gold label for fact verification and $P$ is given by the TAPAS model finetuned on TabFact. 
Larger difference between the predictions for $S$ and $S^\prime_{t}$ indicates the token is more salient.

\subsection{Masked Salient Token Prediction}\label{sec:token_prediction}
% masked salient token prediction as an auxiliary task
% make the salient token more salient
% To exploit the salience of tokens, we introduce a Masked Salient Token Prediction task. The task serves as an auxiliary task in our fact verification system and also an approach for data augmentation.

% salient -> ability
% multi-task
% model; most salient; mask one

Salient tokens in statements, such as %values from tables
lexemes that appear in table cells, and those referring to aggregations and their results, directly contribute to table-text alignment and reasoning. 
Hence, they are critical to table-based fact verification as shown in \Cref{fig:intro}.
Considering the supervision signals for %table-based fact verification
the verification task are sparse and %insufficient to support learning accurate table-text alignment
not necessarily sufficient to capture fine-grained table-text alignment
and the logical relation, we introduce masked salient token prediction as an auxiliary task.

This task is to predict a masked salient token in an entailed statement given the masked statement and the respective table.
We mask the most salient token in each statement according to the salience score estimated in \Cref{sec:salient}.
The reason to do so is that it is hard to find a general threshold to split tokens in different statements into salient and non-salient groups.
% Besides,masking more than one token may cause to multiple valid answers and may be hard to solve.
% in a non-autoregressive \cite{gu2017non} manner.
%Further, %\Cref{sec:result}
The effectiveness of the salience-aware masking will be further evaluated in \Cref{sec:result}.

Both of table-based fact verification and masked salient token prediction share the same TAPAS encoder and the latter reuses the embedding layer as the language modeling head (i.e. linear layer with weights tied to the input embeddings). 
In this way, all %the parameters of the model for the auxiliary task are shared from the model for the main task.
parameters that are updated for the auxiliary task are shared with those in the main task.
%Our system uses multi-task learning \cite{liu2019multi} to urge the model to learn the abilities that are essential to both tasks.
Both tasks are jointly learned, so that the auxiliary task seeks to provide indirect supervision signals to improve the main task. 
The objective function and training details are described in \Cref{sec:training}.

\subsection{Salience-aware Data Augmentation}
\label{sec:aug}
% soft token replacement over non-salient tokens (P7 in slides)
% 1. mask the least salient token in each statement
% 2. use a masked language model to predict the masked token (for efficiency, choose top k)
% 3. use the probability given by MLM as the weights for augmented data.

To effectively learn a robust and generalized NLI model to verify statements based on tables, one requirement is sufficient training data.
Previous work has explored augmenting data 
% by replacing the entity or value in the statement that is also mentioned in the table or 
by filling in specific statement generation templates with entities or values from the table \cite{eisenschlos2020understanding}.
These selected tokens are always detected as salient tokens by the method described in \Cref{sec:salient} as they are important to fact verification.
% However, the training data for fact verification can be limited, while the statements can be presented in very heterogeneous patterns. 
However, previous works ignored the fact that the statement can be presented in heterogeneous ways, % driven by non-salient tokens (muhao: This claim is too strong, if we say that, we need to give references)
and a %ideal (muhao: strong word, better to avoid)
reliable table-based fact verification model should also be adaptive and robust to heterogeneous statements.
In this context, it is intuitive to consider that the 
% least salient tokens 
non-salient tokens
should not interfere the meaning and evidential support of a statement.
Accordingly, we introduce an efficient probabilistic data augmentation technique that leverages the salience of tokens from the other perspective.

% same meaning (FIG 1)
We augment training data by replacing the least salient token in each statement with reasonable alternatives.
Since we expect non-salient token substitution to cause inconsequential meaning change to the original statement, such automatically generated instances will be augmented into the training data along with the original labels.
Similar to \Cref{sec:token_prediction}, we select the least salient token to augment, because it is hard to find a fixed threshold that works for all statements to %distinguish between salient and non-salient ones.
justify whether each of their tokens is important enough or not.

% Once we have obtained the token-level salience estimation from \Cref{sec:salient},
% the process of augmenting data is conducted in three steps. First, %for each statement, we compute the salience scores of all tokens. 
% based on the assumption that non-salient tokens should have insignificant influence on the verification results, we select $K$ tokens with smallest salience scores. These tokens are then masked. 

% weight
% The process of augmenting data is conducted in three steps.
% based on the assumption that non-salient tokens should have insignificant

% Second, a masked language model, BERT, is used to predict a set of possible actual tokens for the masked tokens. 
% In the last step, these masked tokens will be replaced with the predicted tokens and the newly generated statements are assigned the same verification label as the original statement. 
In detail,
for each human-annotated statement, we mask the least salient token and request a BERT model to provide the top $k$ tokens to fill in the blank.
Each a filled token gives an augmented instance of statement.
BERT is pretrained on large textual corpora with the MLM objective, so its predictions can reflect the real-world language expressions\footnote{We do not use TAPAS for data augmentation because the table is not used as input for masked sentence completion.}.
Considering the top $k$ token substitutions are not equally confident according to the BERT predictions and potential noise in data augmentation, 
we down-weight each augmented data instance in training according to the token prediction probabilities (denoted by $w_{ij}$ for the $j$-th augmented instance derived from the $i$-th original instance). 
Related details are presented shortly in \Cref{sec:training}.
% training on those instances will be downweighted according to the token prediction probabilities.
% The associated probabilities are used as weights during training.

\subsection{Training and Inference}\label{sec:training}
We train the model to jointly conduct the main table-based fact verification task (\Cref{sec:basic_model}) using augmented data described in \Cref{sec:aug} %and the masked salient token prediciton model described in
along with the auxiliary task of masked salient token prediction (\Cref{sec:token_prediction}). 

In detail, there are two learning objectives: the binary classification objective $L_v$ for the main task and the MLM objective $L_m$ for the auxiliary task. 
% Formally, given a statement with $k$ tokens, i.e. $C=[v_1, v_2, \cdots, v_k]$, and its label $y$ (entailed or refuted). Let $v_s$ be the token with the highest salience score, and $v_s^\prime$ be the corresponding token predicted by the masked token prediction model. 
% Formally, given the gold label $y_i$ (1 for entailed and 0 for refuted), predicted score $y'_i \in [0,1]$ and augmented data weight $w_i$
% % for the (table, statement) pair $<T_i, S_i>$, 
% of the $i$-th instance in the dataset of size $N_v$,
% the loss function for table-based fact verification is
% \[
% % L_v = \sum_i w_i \cdot CE(y_i, y'_i)
% L_v = - \sum_{i=1}^{N_v}  w_i \cdot (y_i \log(y'_i) + (1-y_i)\log(1-y'_i)).
% \]
% Formally, 
For fact verification, we denote the gold label of the $i$-th instance in the original dataset as $y_i$ (1 for entailed and 0 for refuted).
With salience-aware data augmentation,
each original instance in the dataset is augmented to $k+1$ instances (including itself).
%Each augmented instance is assigned a weight $w_{ij}$ as described in \Cref{sec:aug} ($w_{i0}=1$ for the original instance).
The training instances are also assigned with the probability-based training weight $w_{ij}$ as described in \Cref{sec:aug} ($w_{i0}=1$ for the original instance).
Then, given the model prediction $p_{ij} \in [0,1]$ on each instance,
% , predicted score $y'_i \in [0,1]$ of the $i$-th instance in the original dataset, and the augmented data weight $w_{ij}$ and prediction score $y_{ij}$ of the $j$-th augmented instance derived from the $i$-th original instance,
the loss function is defined as the following weighted cross-entropy, where $N_v$ is the number of instances in the original dataset:
\[
L_v =  - \sum_{i=1}^{N_v} \sum_{j=0}^{k} w_{ij} (y_i \log(p_{ij}) +  (1-y_i) \log(1-p_{ij})).
\]
% \begin{equation*}
% L_v = - \sum_{i=1}^{N_v} ( (y_i \log(y'_i) + (1-y_i)\log(1-y'_i)) + \\
% \sum_{j=1}^{k} w_i^j \cdot (y_i \log(y_i^j) + (1-y_i)\log(1-y_i^j)) ).
% \end{equation*}
% \begin{equation*}
% \begin{split}
% L_v = & - \sum_{i=1}^{N_v} (y_i (\log(y'_i) + \sum_{j=1}^{k} w_{ij} \log(y_{ij}) )  \\
%  & + (1-y_i) (\log(1-y'_i) + \sum_{j=1}^{k} w_{ij} \log(1-y_{ij}) ) )
% \end{split}
% \end{equation*}
% where $k$ is the number of augmented instances derived from each original instance and $N_v$ is the number of instances in the original dataset.
For the auxiliary task, given the gold label $y_i^j$ (1 for the target token, 0 for other tokens) and % the predicted probability 
model outputs $p_i^j$ of each candidate token $c_j \in V$ for the $i$-th instance, the loss function is defined as below, where $N_m$ is the number of all entailed statements in the dataset:
\[
L_m = - \sum_{i=1}^{N_m} \sum^{|V|}_{j=1} y_i^j  \log(p_i^j) .
\] 
The overall learning objective is to optimize the following joint loss, where $\alpha$ is a coefficient to balance between the two task objectives:
\[
L =  \frac{\alpha}{N_v k}  L_v + \frac{(1 - \alpha)}{N_m}  L_m .
\] 
In inference, given a statement and a table, we use the prediction head of fact verification independently and perform the verification without augmenting the test data, 
%as NLI with the table premise and the statement hypothesis. This follows the details in \Cref{sec:basic_model}. 
following the details in \Cref{sec:basic_model}. 

% Let $v_s$ be the token with the highest salience score, and $v_s^\prime$ be the corresponding token predicted by the masked token prediction model. 
% $L_M$ and $L_v$ is defined as
% \[L_{M} = CE(v, v^\prime)\]
% and
% \[L_v = CE(f(C), y)\]
% where $f(C)$ represents the prediction.
% The overall loss function is
% \[L = w \cdot(L_M + L_v)\] where $w$ is the weight of the statement. If the statement is generated from the data augmentation model, $w$ is the probability associated with it, otherwise, $w=1$.

% \[
%     L_e = 
% \]
% and 
% \[
%     L_m = 
% \].
% The final loss function $L$ is a linear combination between $L_e$ and $L_m$

% \[
%     L = \alpha L_e + (1 - \alpha) L_m
% \]
% where $\alpha$ is a hyperparameter showing the weights of $L_e$ and $L_m$. We set $K$ to be $1$ and $\alpha$ to be $0.5$ which means the two parts of the model contribute equally.

%% file: sections/4_experiment.tex
%Statistics of the TabFact dataset are listed in \Cref{table:dataset}. Each table comes along with 2 to 20 statements, and consists of 14 rows and 5 columns in average.

\section{Experiment}
In this section, we conduct experiments on the TabFact dataset. We first introduce the dataset, a series of recent baselines and details of our method (\Cref{sec:exp_set}). Then we show the overall performance and ablation results (\Cref{sec:result}). We also provide case studies for in-depth analysis (\Cref{sec:case}).

\subsection{Experimental Settings}\label{sec:exp_set}

\stitle{Dataset and Evaluation}
We evaluate our model on the TabFact benchmark \cite{chen2019tabfact} that is widely used by studies on this task\footnote{https://tabfact.github.io/}.
The dataset contains $118,275$ statements and  $16,573$ tables.
Each table thereof comes along with 2 to 20 statements, and consists of 14 rows and 5 columns in average.
Each statement is paired with a table and is labeled as entailed or refuted by information in the table.
We use the originally released train, validation and test splits for evaluation, for which the statistics are listed in \Cref{table:dataset}.
Tables in these splits do not have overlaps.
%Specifically, statements are also categorized into simple or complex ones according to the verification difficulty, and a small test set is sampled and evaluated by human annotators.
Specifically, statements in the test split are further labeled into simple or complex categories according to their verification difficulty.
Additionally, a small subset of the test split is used to compare machine performance and human performance.
Being consistent with previous studies \cite{chen2019tabfact, zhong2020logicalfactchecker, yang2020program, eisenschlos2020understanding},
we report the model performance on the validation and test splits, two of the difficulty-specific subsets, as well as the small subset with human performance, and use accuracy as the evaluation metric.

\begin{table}[t]
\centering
\small
\begin{tabular}{ccc}
\toprule%[2pt]
Split & \#Statement & \# Table  \\ \midrule
Train & 92,283 & 13,182 \\
Validation & 12,792 & 1,696 \\
Test & 12,799 & 1,695 \\ \midrule
Simple & 50,244 & 9,189 \\
Complex & 68,031 & 7,392 \\
\bottomrule%[2pt]
\end{tabular}
\caption{Statistics of the TabFact dataset.}
\label{table:dataset}
\end{table}

\begin{table*}[t]
\centering
{
\small
\begin{tabular}{l|p{1.5cm}<{\centering} p{1.5cm}<{\centering} ccc}
\toprule
Model & Val & Test & Test (simple) & Test (complex) & Small Test Set \\ 
\midrule
Human Performance & - & - & - & - & 92.1 \\
\midrule
LPA & 65.2 & 65.0 & 78.4 & 58.5 & 68.6 \\
LogicalFactChecker &71.8& 71.7& 85.4& 65.1& 74.3\\
HeterTFV & 72.5 & 72.3 & 85.9 & 65.7 & 74.2 \\
ProgVGAT& 74.9 & 74.4& 88.3& 67.6& 76.2\\ \midrule
Table-BERT &66.1&65.1&79.1&58.2&68.1 \\
TAPAS \cite{dong2021structural} & - & 76.0 & 89.0 & 69.8 & - \\
TAPAS \cite{eisenschlos2020understanding} &81.0 &81.0& 92.3& 75.6& 83.9 \\
\midrule
ours & \textbf{82.7} & \textbf{82.1} & 93.3 & \textbf{76.7} & 84.3 \\
\ \ -- w/o augmented data & 82.4 & 82.1 & 93.4 & 76.6 & \textbf{84.4} \\
\ \ -- w/o auxiliary task & 81.8 & 81.9 & \textbf{93.6} & 76.3 & 84.1 \\
\bottomrule
\end{tabular}
}
\caption{Performance on the official splits of TabFact in terms of verification accuracy (\%).
Baselines are organized into \textit{logical program-driven} (i.e. LPA, LogicalFactChecker, HeterTFV and ProgVGAT) and \textit{non-logical program-driven} (i.e. Table-BERT and TAPAS). Human performance is reported by \citet{chen2019tabfact}.
}
%\vspace{-1em}
\label{table:result}
\end{table*} 

\stitle{Baselines}
We compare our system with the following competitive baselines:
\begin{itemize}[leftmargin=1em]
\setlength\itemsep{0.2em}
\item Latent Program Algorithm (LPA) \cite{chen2019tabfact} synthesizes logical programs based on the given statement and table, executes programs to return bool labels, and aggregates the results according to the confidence score of each program assigned by a Transformer-based model. 

\item LogicalFactChecker \cite{zhong2020logicalfactchecker} captures token-level semantic interaction between a statement, a table and a derived program using BERT with graph-based masking. Logical semantics of each program is captured with neural module networks \cite{andreas2016learning}. 
% The final prediction is based on semantic representations of both levels.
% obtain representations of the statement, table and program by graph representation learning, and model the logical operations by different neural networks.
% utilizes inherent structures of programs to prune irrelevant information in evidence tables and modularize symbolic operations with module networks.

\item HeterTFV \cite{shi2020learn} constructs a heterogeneous graph to incorporate the statement, the table and the program, and applies a heterogeneous graph attention network to capture both linguistic and symbolic information.

\item ProgVGAT \cite{yang2020program} generates a program and verbalize the execution progress as evidences. The system applies a graph attention network \cite{velivckovic2017graph} to capture the execution graph, the table and statement.
% Then the system build a graph based on the program structure with an additional node representing the table and statement, and applies a graph attention network to make predictions.

\item Table-BERT \cite{chen2019tabfact} applies BERT for NLI taking a statement as the hypothesis and a linearized table as the premise.

\item TAPAS \cite{herzig2020tapas} is a Transformer-based model pre-trained on textual and tabular data. \citet{dong2021structural} and \citet{eisenschlos2020understanding} have formulated table-based fact verification as an NLI task, and applied TAPAS with task-specific intermediate pretraining. The latter one achieves the current SOTA performance on TabFact.

\end{itemize}

\stitle{Model Configurations}
% general
Our system also adopts the officially released TAPAS-Large model, which applies intermediate pre-training and is fine-tuned on TabFact, as our base model\footnote{https://github.com/google-research/tapas}.
Following \citet{eisenschlos2020understanding}, we set the max input length to $512$.
We use $10,000$ training steps, and optimize the learning objective with an AdamW optimizer \cite{loshchilov2018decoupled} which sets the learning rate to $5e^{-5}$, a batch size of $32$ and a warmup ratio of $0.1$.
All hyper-parameters are decided according to the validation performance.
% mlm
For multi-task learning, we set the coefficient between two losses $\alpha$ to $0.5$.
% aug
For data augmentation, we use the uncased BERT-Large model as the MLM.
For computational efficiency, we select the top $k=3$ predictions for probabilistic data augmentation.
% Reproducibility Checklist
% It takes around 36 hours to train the full model. 

\subsection{Results}\label{sec:result}

\stitle{Overall Performance}
% baselines
% ours against the key baseline
%   val and test
%   human performance
\Cref{table:result} presents the results of different verification models. 
Among the baseline methods, TAPAS with task-specific intermediate pretraining demonstrates %SOTA 
the best performance. %over other baselines. 
It implies that explicit logical programs is not a necessity for reasoning between the table and the statement.
We observe that our system outperforms the best baseline with 2.1\% relative improvement on the validation set and 1.4\% relative improvement on the test set in terms of accuracy.
It is noteworthy that, our system applies the same backbone model and pretraining process as the previous best method, so that all the improvements are %from
attributed to the salience-aware learning strategies.
Besides, our system reduces the gap between machine performance and human performance on the small test set to 7.8\%.
These experimental results verify our hypothesis that masked salient token prediction and salience-aware data augmentation are conducive to table-based fact verification.

% \begin{figure}[t]
% \centering
% \includegraphics[scale=0.36]{figures/ablation_2d.png}
% \caption{
% Ablation results on the validation and test set. 
% The middle group is the results for the default system.
% For each group, the left bar is accuracy on the validation set and the right one is for the test set.
% }
% \label{fig:ablation}
% %\vspace{-1.5em}
% \end{figure}

\begin{table}[h]
\small
\centering
\begin{tabular}{c|l|cc}
\toprule
 & Strategy & Val & Test   \\ \midrule
\multirow{2}{*}{Masking} & Random  & 82.1 & 81.9 \\
& Salient  & \textbf{82.4} & \textbf{82.1} \\ \midrule
\multirow{2}{*}{Augmentation} & Uniform  & 81.5 & 81.3 \\
& Probabilistic  & \textbf{81.8} & \textbf{81.9} \\
\bottomrule
\end{tabular}
\caption{
Ablation results for masking strategy and augmentation strategy.
To avoid co-effects, we conduct experiments on  masking (or augmentation) strategy  without using augmented data (or auxiliary task).}
\label{table:ablation}
% \vspace{-1em}
\end{table}

\begin{table*}[t]
\small
\centering
\begin{tabular}{  m{12cm} | m{1.6 cm} m{1 cm} } \toprule
% \makecell[c]{Table Caption} & 
\makecell[c]{Token Salience Estimation} & \multicolumn{2}{c}{Augmentation}
\\ \midrule
% F.C. United of Manchester
% &
{\setlength{\fboxrule}{1pt}
\fcolorbox{white}{orange!5}{The} 
\fcolorbox{white}{orange!10}{file}
\fcolorbox{white}{orange!20}{format} 
\fcolorbox{white}{orange!30}{mobipocket} 
\fcolorbox{red}{orange!2}{comes}
\fcolorbox{white}{orange!5}{with}
\fcolorbox{white}{orange!30}{all}
\fcolorbox{white}{orange!50}{three}
\fcolorbox{blue}{orange!80}{supports}
\fcolorbox{white}{orange!5}{.}
}
% the file format mobipocket comes with all three supports.
&
\makecell[l]{
works \\
worked \\
compatible \\
}
& 
\makecell[l]{
0.615 \\
0.051 \\
0.029
} 
\\  \midrule
% Miss Mundo Dominicana 2007
% &
{\setlength{\fboxrule}{1pt}
\fcolorbox{white}{orange!5}{The} 
\fcolorbox{red}{orange!2}{player}
\fcolorbox{white}{orange!5}{from} 
\fcolorbox{white}{orange!5}{the} 
\fcolorbox{blue}{orange!80}{Santiago}
\fcolorbox{white}{orange!40}{province}
\fcolorbox{white}{orange!5}{lives}
\fcolorbox{white}{orange!5}{in}
\fcolorbox{white}{orange!5}{the}
\fcolorbox{white}{orange!40}{city} 
\fcolorbox{white}{orange!20}{navarrete}
\fcolorbox{white}{orange!5}{.}
}
% The player from the santiago province lives in the city navarrete
&
\makecell[l]{
population \\
people \\
one 
}
& 
\makecell[l]{
0.352\\
0.184 \\
0.035
}
\\ \midrule
% Mixed martial arts record
% &
{\setlength{\fboxrule}{1pt}
\fcolorbox{white}{orange!78}{Canton} 
\fcolorbox{white}{orange!2}{,}
\fcolorbox{white}{orange!5}{Ohio} 
\fcolorbox{white}{orange!5}{was} 
\fcolorbox{white}{orange!5}{the}
\fcolorbox{white}{orange!5}{location}
\fcolorbox{red}{orange!2}{for}
\fcolorbox{white}{orange!2}{the}
\fcolorbox{white}{orange!5}{event}
\fcolorbox{white}{orange!5}{,} 
\fcolorbox{white}{orange!40}{fightfest}
\fcolorbox{white}{orange!5}{2}
\fcolorbox{white}{orange!5}{,}
\fcolorbox{white}{orange!5}{which}
\fcolorbox{white}{orange!5}{lasted}
\fcolorbox{white}{orange!5}{only}
\fcolorbox{blue}{orange!80}{3}
\fcolorbox{white}{orange!10}{rounds}
\fcolorbox{white}{orange!5}{.}
}
% canton , ohio was the location for the event , fightfest 2 , which lasted only 3 rounds
&
\makecell[l]{
of \\
to \\
in \\
}
& 
\makecell[l]{
0.709 \\
0.001 \\
0.001
}
% \\ \midrule
% % F.C. United of Manchester
% % &
% {\setlength{\fboxrule}{1pt}
% \fcolorbox{white}{orange!5}{The} 
% \fcolorbox{red}{orange!2}{league}
% \fcolorbox{white}{orange!20}{contested} 
% \fcolorbox{white}{orange!20}{in} 
% \fcolorbox{white}{orange!20}{2011}
% \fcolorbox{white}{orange!20}{-}
% \fcolorbox{white}{orange!20}{12}
% \fcolorbox{white}{orange!20}{,}
% \fcolorbox{white}{orange!5}{and}
% \fcolorbox{white}{orange!20}{level} 
% \fcolorbox{blue}{orange!60}{7}
% \fcolorbox{white}{orange!5}{,}
% \fcolorbox{white}{orange!30}{is}
% \fcolorbox{white}{orange!5}{the}
% \fcolorbox{white}{orange!20}{northern}
% \fcolorbox{white}{orange!5}{premier}
% \fcolorbox{white}{orange!2}{league}
% \fcolorbox{white}{orange!20}{premier}
% \fcolorbox{white}{orange!20}{division}
% \fcolorbox{white}{orange!5}{.}
% }
% % the league contested in 2011 - 12 , and level 7 , is the northern premier league premier division
% &
% \makecell[l]{
% division \\
% competition \\
% tier \\
% }
% & 
% \makecell[l]{
% 0.446 \\
% 0.163 \\
% 0.064
% }
\\ \bottomrule
\end{tabular}
\caption{
Examples of salience estimation and data augmentation. 
Darker background indicates %a smaller salience score. 
more salience.
\fcolorbox{blue}{white}{Blue} rectangles mark the targeted most salient tokens in masked salient token prediction.
\fcolorbox{red}{white}{Red} rectangles mark the least salient tokens that are to be %replaced by the tokens with the corresponding weights in the second column for data augmentation.
substituted by the augmentation tokens, for which weights are listed.
}
\label{table:case}
%\vspace{-1em}
\end{table*}

% south melbourne 's home side score is 12.13 (85)

\stitle{Effect of Masked Salient Token Prediction}
% w/o augmented data (setting + result + conclusion)
% salient v.s. random
The performance of the base model with masked salient token prediction is marked as ``{w/o augmented data}'' in \Cref{table:result}.
The auxiliary task solely brings along 1.7\% relative improvement on the validation set and 1.4\% relative improvement on the test set.
This demonstrates that the indirect supervision brought by the auxiliary task can directly benefit the main task training.
\Cref{table:ablation} compares salient masking and random masking 
% (the two bars at the beginning and end of the arrow \texttt{random masking}
for the auxiliary task.
For fair comparison, we mask one token in each entailed statement for both strategies.
The results show that salient masking reduces error rate on the validation set by relative 1.7\% (and by relative 1.1\% on the test set) in comparison with random masking.
This is not surprising since random masking may mask non-salient tokens which %can not provide indirect supervision signals for table-text alignment and logical reasoning.
are not decisive for table-text alignment and logical inference.

\stitle{Effect of Salience-aware Data Augmentation}
% w/o auxiliary task
% probabilistic v.s. uniform
The performance of the base model with salient-aware data augmentation is marked as ``w/o auxiliary task'' in \Cref{table:result}.
The data augmentation independently brings 1.0\% relative improvement on the validation set and 1.1\% relative improvement on the test set.
The results demonstrate that table-based fact verification requires abundant training data and verify the effectiveness of the proposed data augmentation strategy.
\Cref{table:ablation} compares probabilistic weights and uniform weights.
% (the two groups under the edge \texttt{uniform data weight}) for augmented data.
The results show that probabilistic data augmentation reduces error rate on the validation set by 1.6\% relatively (and by 3.2\% relatively on the test set) in comparison with uniform data augmentation.
This observation is reasonable because the augmented data are not equally confident according to the MLM predictions.
Moreover, the predicted probabilities from the pretrained language model %can reflect real-world distribution in some degree.
correlate with real-world distribution of English language.

\stitle{Performance on Simple and Complex Instances}
% better on both simple and complex
% Masked Salient Token Prediction -> better on complex
% Salience-aware Data Augmentation -> better on simple
We further compare the performance of baselines and variants of our system on two groups of test instances labeled with different verification difficulties. 
Our system outperforms all the baselines on both simple and complex instances with at least 1.0\% absolute improvement.
Ablation results in \Cref{table:result} also show that the auxiliary task improves the base model more on complex instances while data augmentation improves the base model more on simple instances.
These results are consistent with the features of the two salience-aware learning strategies.
Masked salient token prediction seeks to enhance the model %ability of
to capture table-text alignment and %logical reasoning 
the underlying logical relations,
so that complex instances %which need complicated reasoning benefit more.
requiring more complicated reasoning gain more benefits.
Salience-aware data augmentation seeks to 
% enhance model robustness 
% to diverse expressions 
augment statements
by simply replacing non-salient tokens.
% replaces non-salient tokens in each statement.
This strategy increases the training data but does not augment the implicit logical form covered by the dataset so that the improvement on complex instances is not as significant as that on simple instances.

% \subsection{Ablation Study}

% \stitle{Effect of Multi-task Learning}

% \stitle{Effect of Salient Masking}

% \stitle{Effect of Data Augmentation}

\subsection{Case Study}
\label{sec:case}
% salient tokens can be entities, numbers, higher/largest/...
% close-world, salient token based on the given table, some tokens are not ambiguous even when masked. that is why 'player' can be non-salient
% augmented data are helpful. extend the expressions
% weakness: multi-token words, can not detect. (e.g. '3290350', 'fightfest 2'). It further reflects a potential problem of existing verification models.
% augmented data can be noisy (e.g. 'population')

% general
We present a case study with three representative examples to illustrate salience estimation and data augmentation in \Cref{table:case}.
% salient
The detected salient tokens can be entities and numeric values from the table, tokens indicating %logical operations over the table
relations, and the results of logical operations.
% non-salient
Non-salient tokens can be common nouns, verbs, prepositions and so on.
These tokens are detected as non-salient because they %do not provide specific information according to the given table.
are not closely associated with facts in the given table.
For example, the table in the second example is about the residence of different athletes, so ``\texttt{player}'' in the statement %is a general term.
may be substituted to related terms without interfering the verification result.
% augmentation
%Besides, the augmented data, though exists some noise, effectively improves the statement diversity.
% risk: multi-word entity
It is noteworthy that entities consisting of multiple words tend to have relatively small salience scores for some parts.
It may be due to that verification models can identify the corresponding cell by part of the entity.
But it also raises the risk of incorrect verification or %being attacked
polluted data augmentation when modifying a part of a multi-word entity.

%% file: sections/5_conclusion.tex
\section{Conclusion}

In this paper, we proposed a novel system for table-based fact verification.
Our system employs salience-aware learning %to enhance the basic model TAPAS for capturing fine-grained connections and supporting the reasoning between statements and tables and improve the model robustness to diverse statements.
and introduce complementary supervision signals by leveraging both salience and non-salient tokens from different perspectives. 
The system consists of three key techniques, including probing-based salience estimation, masked salient token prediction and salience-aware data augmentation.
Experiments on the TabFact benchmark show that our system leads to significant improvements over the current SOTA systems.
For future work, we plan to extend salience-aware learning to other NLU tasks, including NLI \cite{bowman2015large, williams2018broad} and Tabular QA \cite{sun2016table,chen2020hybridqa}.
%\cite{feldman2013techniques, maas2011learning}.
Applying the idea of salience estimation to NLG tasks, such as controlled table-to-text generation \cite{parikh2020totto} and paraphrasing \cite{iyyer2018adversarial, huang2021generating}, is another meaningful direction. 

%% file: sections/6_ethics.tex
\section*{Ethical Consideration}

This work does not present any direct societal consequence. 
The proposed work seeks to develop a salience-aware learning framework for fact verification using tabular data as evidence. 
We believe this leads to intellectual merits that benefit claim and statement verification for Web corpora, as well as detection of misinformation. 
It potentially also has broad impacts %since the tackled issues also widely exists in tasks of other areas.
for NLU and NLG tasks where tables serve as a medium of knowledge sources.
The experiments are conducted on a widely-used open benchmark.

The goal of this research topic is to help identify misinformation, which seeks to benefit societal fairness.
While we treat tables as reliable sources of evidences like relevant studies do, we do not hypothesize that the populated information by Web users in tables is not completely free of societal bias. We believe this is a meaningful research direction for further exploration. While not being explicitly studied in this work, the incorporation of salience-aware inference could be a way to control or mitigate societal biases.

\section*{Acknowledgement}

We appreciate the anonymous reviewers for their insightful comments.

This research is supported by the Defense Advanced Research Projects Agency (DARPA) and the Air Force Research Laboratory (AFRL) under contract number FA8650-17-C-7715, by the DARPA MCS program under Contract No. N660011924033 with the United States Office Of Naval Research, and by the National Science Foundation of United States Grant IIS 2105329.
The U.S. Government is authorized to reproduce and distribute reprints for Governmental purposes notwithstanding any copyright notation thereon. The views and conclusions contained herein are those of the authors and should not be interpreted as necessarily representing the official policies or endorsements, either expressed or implied, of the U.S. Government.